\pdfoutput=1

\documentclass[11pt]{article}

\usepackage[final]{acl}

\usepackage{times}
\usepackage{latexsym}

\usepackage[T1]{fontenc}

\usepackage[utf8]{inputenc}

\usepackage{microtype}

\usepackage{inconsolata}

\usepackage{graphicx}

\usepackage{amsmath,amssymb,amsfonts,amsthm}

\usepackage{hyperref}
\usepackage{geometry}

\usepackage{enumitem}
\usepackage{titlesec}
\usepackage{stmaryrd}
\usepackage{mathrsfs}
\usepackage{tikz}
\usepackage{txfonts}
\usepackage{hyperref}

\usepackage{lingstyle}

\usepackage{xcolor}
\usepackage{tcolorbox}
\usepackage{dialogue}
\usepackage{algorithm}
\usepackage{algpseudocode}
\usepackage{booktabs}
\usepackage{subcaption}

\title{Dynamic Epistemic Friction in Dialogue}

\author{
  Timothy Obiso$^{1}$ \quad
  Kenneth Lai$^{1}$ \quad
  Abhijnan Nath$^{2}$ \\
  {\bf Nikhil Krishnaswamy}$^{2}$ \quad
  {\bf James Pustejovsky}$^{1}$ \\
  $^{1}$Brandeis University, Waltham, MA USA \\
  $^{2}$Colorado State University, Fort Collins, CO USA \\
  \texttt{\{timothyobiso, klai12, jamesp\}@brandeis.edu} \\
  \texttt{\{abhijnan.nath, nkrishna\}@colostate.edu}
}

\begin{document}
\maketitle

\begin{abstract}
Recent developments in aligning Large Language Models (LLMs) with human preferences have significantly enhanced their utility in human-AI collaborative scenarios. However, such approaches often neglect the critical role of "epistemic friction," or the inherent resistance encountered when updating beliefs in response to new, conflicting, or ambiguous information. In this paper, we define {\it dynamic epistemic friction} as the resistance to epistemic integration, characterized by the misalignment between an agent's current belief state and new propositions supported by external evidence. We position this within the framework of Dynamic Epistemic Logic \cite{van2011logical}, where friction emerges as nontrivial belief-revision during the interaction. We then present analyses from a situated collaborative task that demonstrate how this model of epistemic friction can effectively predict belief updates in dialogues, and we subsequently discuss how the model of belief alignment as a measure of epistemic resistance or friction can naturally be made more sophisticated to accommodate the complexities of real-world dialogue scenarios.

\end{abstract}

\section{Introduction}

In cooperative, well-grounded conversations, the exchange of information often appears straightforward. 
Participants typically assume that updates to one another’s beliefs will be smooth and consistent with mutual common ground. 
A listener hears a speaker’s assertion and, assuming trust and shared context, incorporates it into their beliefs with minimal hesitation. However, in many situations---including disputes and strategic deception, but also innocent misalignment in good-faith collaborations---new information generates \emph{resistance} to belief revision. In these cases, not all updates fit so neatly. Sometimes, new information conflicts with the listener’s prior understanding, challenges their assumptions, or signals a hidden agenda. Here, the process of updating belief states is not "frictionless." Instead, the listener encounters a kind of "resistance" to easy assimilation, a phenomenon we call \emph{epistemic friction}.

Friction in conversational updates reflects an underlying complexity in how we process and accommodate new information, while pointing to deeper inferential processes within the participants' epistemic state. Understanding friction can help us identify when a speaker might be deceptive, when a conversation is strategically misaligned, or when a seemingly simple statement actually encodes a more complex epistemic move. In short, friction offers insight into the subtle interplay between logical inference, pragmatic reasoning, and the architecture of cognitive representations.

In physical systems, friction is a force that resists motion. By analogy, epistemic friction is a resistance to the smooth "motion" of belief revision. This resistance might be epistemically beneficial—encouraging the listener to scrutinize the new information more carefully, or to consider alternative explanations. It might also expose underlying strategic interests, deceptive behavior, or complexities in the conceptual structure of what is being communicated.  Here, we explore frictive interactions in terms of evidence-based dynamic epistemic logic (DEL;~\citet{van2011logical}), a well-established logical framework for modeling belief updates, as recently explored in \citep{khebour2024common}.  

We introduce a vector-based modeling approach, drawing on Holographic Reduced Representations (HRR) \cite{plate1995holographic,luo2018towards} and related vector symbolic architectures \cite{kanerva1988sparse}. This approach treats agents' belief states and propositions as high-dimensional vectors, allowing geometric notions like orthogonality and angle to characterize the friction that arises when assimilating new information. By bridging the gap between symbolic logic and geometric intuition, this model provides a novel perspective on the cognitive and communicative processes underlying conversation.

Finally, we provide case studies from a situated collaborative task that demonstrate how this model of epistemic friction can be used to create a straightforward vectorization of task-relevant propositionalized beliefs and their subsequent updates in the face of new interlocutor assertions. Our analyses demonstrate the utility of epistemic friction in both modeling dialogues and in human-AI interactions, and we subsequently discuss how the model of belief alignment as a measure of epistemic resistance or friction can naturally be made more sophisticated to accommodate the complexities of real-world dialogue scenarios.

\section{Related Work}
\label{sec:related}

Epistemic friction is clearly related to the classic notions of miscommunication and misalignment of common ground in conversation \cite{grice1975logic,asher2003common,stalnaker2002common,
traum2003information}. The concept of common ground refers to the  set of shared beliefs among participants in a Human-Human interaction (HHI) ~\cite{markowskaformal,traum1994computational,hadley_review_2022}, as well as HCI \cite{krishnaswamy2020formal,pustejovsky2021embodied,ohmer2022emergence} and HRI interactions \cite{kruijff2010situated,fischer2011people,scheutz2011toward}. 
When common ground is lacking or divergent, interlocutors experience misunderstandings or must exert an effort to clarify and realign their beliefs \cite{clark1986referring}. 
Such effortful moments are essentially points of friction. Although friction is typically seen as something to overcome or mitigate in dialogue \cite{brown-etal-2003-reducing,hunter2018formal}, friction can also play a beneficial role in the interaction \cite{chen2024exploring}. 

 In Dynamic Epistemic Logic (DEL), degrees of evidence (or strength of belief)  towards  a proposition, can be seen as correlated to the friction that an agent has towards a public proposition \cite{van2007dynamic,van2015dynamic,van2011logical}.  Similarly, in argumentation theory, friction can be seen as analogous to the degree of acceptance or rejection of beliefs in an argument \cite{baumann2015agm,hunter2020epistemic}.  From this perspective, friction is not only about the endpoint of belief revision but about the trajectory: how beliefs resist, adapt, or transform as agents encounter a continuous stream of arguments and evidence.

Beyond logical and probabilistic formalisms, researchers have explored vector-space representations of propositions within distributional models  \cite{baroni2013composition,
boleda2020distributional,
lenci2023distributional}, 
as well as hyperdimensional  models 
\cite{plate1995holographic,kanerva1988sparse,ginzburg2024,obiso2024holographic}. Within the areas  of dialogue and multiparty interactions, vector models of propositional content have been employed in the service of tracking  common ground \cite{khebour2024common,zhu2024modeling,palmer2024speech}.

 \section{Epistemic Friction in Communication}
 \label{sec:epistemic-friction}



A core assumption in many theories of discourse, ranging from Grice’s cooperative principle \cite{grice1975logic} to Stalnaker’s common ground framework \citep{stalnaker2002common}, is that participants in a conversation share a basis of mutual knowledge and strive for coherence. However, in many situations new information generates \emph{resistance} to belief revision. 
These situations may include adversarial or cooperative-competitive situations such as disputes or strategic deception \cite{niculae2015linguistic}, but also ordinary good-faith collaboration. In these cases, a
listener hears a speaker’s assertion and, assuming trust and shared context, incorporates it into their beliefs with minimal hesitation. Nevertheless, not all updates fit so neatly. Sometimes, new information conflicts with the listener’s prior understanding, challenges their assumptions, signals a hidden agenda, 
or this misunderstanding or misremembering mutates the information the listener believes they are incorporating
. In these cases, the process of updating belief states is not "frictionless." Instead, the listener encounters a kind of "resistance" to easy assimilation, a phenomenon we call \emph{epistemic friction}. In the context of a constantly updating dialgoue, we call this phenomenon {\it dynamic epistemic friction}.

In DEL, we use a standard modal model, $M = (W, \{R_a\}_{a \in \mathcal{A}}, V)$, where:
\enumsentence{
a. $W$ is a set of possible worlds;
\\
b. $R_a$ is the accessibility relation for agent $a$,
\\
c. $V$ is a valuation function assigning truth conditions to atomic propositions.
}
\noindent
Knowledge or belief operators ($B_a \varphi$) are evaluated by requiring $\varphi$ to hold in all $R_a$-accessible worlds.
DEL captures belief change by product updates with event models \cite{bolander2014seeing}. Formally, an event model $\mathcal{E} = (E, \{R_a^E\}, \mathrm{pre})$ is combined with $M$ as in (\ref{eventmodel}), where $\otimes$ denotes the product update:
\enumsentence{
a. $ M \otimes \mathcal{E} = (W \times E,\;\{R_a^\otimes\},\;V^\otimes) $ \\
b. where $(w, e) R_a^\otimes (w', e')$ iff $w R_a w'$, $e R_a^E e'$, and $M,w \models pre(e)$ and $M,w' \models pre(e')$.
\label{eventmodel}
}
\noindent 
If an event is \emph{public}, each agent’s belief set typically refines (or filters) to those worlds consistent with the event's precondition. Usually, we assume that all agents smoothly integrate the new proposition. But if the proposition conflicts strongly with the agent's prior beliefs, friction ensues.

We say friction occurs when an agent’s newly updated beliefs \emph{cannot} be derived by a simple monotonic restriction of the old ones. Formally, consider an agent $a$ with old beliefs $B_a^{old}$, updated by $\psi$ to $B_a^{new}$. 
Alignment is quantified by checking how trivially \(\psi\) is entailed by \(B_a\). Friction occurs when updates require epistemic revision, formally:
\enumsentence{
$B^{new} \not\subseteq B^{old} \cup \{\psi \mid B^{old} \vdash \psi\}$
}
\noindent
Conversely, a lack of friction corresponds to minimal cognitive effort in integrating new propositions.

\citet{khebour2024common} introduce the framework of \emph{evidence-based DEL}, in which common ground is structured into:
\vspace*{-2mm}
\eenumsentence{
\item\textbf{QBank} (Questions Under Discussion): Propositions requiring evaluation.
\vspace*{-2mm}
\item \textbf{EBank} (Evidence Bank): Propositions with supporting evidence.
\vspace*{-2mm}
\item \textbf{FBank} (Fact Bank): Propositions accepted as true.
}
\vspace*{-2mm}
\noindent In this framework, one tracks how propositions move from the \emph{Question Bank (QBank)} to the \emph{Evidence Bank (EBank)} and eventually to \emph{Fact Bank (FBank)} when evidence is deemed sufficient \cite{ginzburg1996dynamics}. When new evidence $[E]\varphi$ enters, high friction signals that $\varphi$ is \emph{misaligned} with the agent’s prior or insufficiently supported. As more supporting evidence accumulates, friction reduces.

How can we infer the beliefs $B_a$ of an agent $a$? Following \citet{bolander2014seeing} and \citet{zhu2024modeling}, we can obtain evidence for what an agent believes from what they do, say, or perceive, formalized in the following axioms:

\eenumsentence{
\item 
{\bf Acting is Believing:} $DO_a \varphi \rightarrow B_a \varphi$ (you believe your own actions)
\\ As an agent participant in an event, you believe it has happened. 
\item 
{\bf Saying is Believing:} $SAY_a \varphi \rightarrow B_a \varphi$ (you believe what you say) \\
As actor of a declarative speech act, you believe the proposition you express. 
\item 
{\bf Seeing is Believing:} $SEE_a \varphi \rightarrow B_a \varphi$ (you believe what you see) \\
As witness to a situation or event, you believe it to have occurred.

}
\label{assumptions}

 \section{Epistemic Alignment}
\label{sec:alignment}
 
Suppose an agent $a$ has a belief state $B_a\subseteq W$, where $W$ is the set of possible worlds that the agent considers viable. Let $\{w \in B_a \mid w \models \varphi\}$ be the subset of worlds in which $\varphi$ holds, and let $E$ be some set of "evidence worlds". In the context of modal logic, $B_a$ functions as a modal operator; in the context of alignment and misalignment, $B_a$ is interpreted as a predefined set.
A straightforward way to define epistemic alignment is to define what fraction of $a$’s currently possible worlds also satisfy $\varphi$ (and are consistent with the evidence $E$). That is: 
\vspace*{-2mm}
$$
\mbox{alignment}(\varphi, B_a, E) = \frac{|\{w \in B_a \mid w \models \varphi\} \cap E|}{|B_a|}
$$
\vspace*{-2mm}

 If almost all of $B_a$ already support $\varphi$, then alignment $\approx 1$, so friction is low.  If few or none of the worlds in $B_a$ satisfy $\varphi$, alignment $\approx 0$, so friction is high. One can define "consistent with $E$" in many ways (e.g., requiring each $w\in B_a$ to also satisfy whatever constraints the evidence imposes). The key idea is that alignment measures how large the overlap is between $\varphi$ and the agent’s current doxastic possibilities, modulated by the evidence.

If we consider the propositional content as dense vector encodings, then 
%
we can define $\mathbf{v}_{B_a}$ to be the vector encoding agent $a$’s overall belief state,  
$\mathbf{v}_\varphi$ to be a vector encoding the proposition $\varphi$,  
and  $\mathbf{v}_E$ to be a vector encoding relevant evidence $E$.
A natural strategy is to use cosine similarity due to its prevalence in HRR \cite{plate1995holographic}. However, the choice of similarity function may depend on the algebras or symbolic logic used to represent propositions in a given system \cite{Kleyko_2022, Kleyko_2023}. This function should also be chosen based on the way propositional content is vectorized in the propositionalized vector.

In our case, a simple encoding treats "$\varphi + E$" as the combined proposition‐plus‐evidence vector,  measuring its similarity to the agent’s belief vector:
$$
\text{alignment}(\varphi, B_a, E) = \text{CosSim}(\mathbf{v}_{B_a}, \mathbf{v}_\varphi + \mathbf{v}_E)
$$

\noindent
A large positive dot product indicates high alignment, while a near-zero or negative dot product indicates strong orthogonality or conflict, meaning the agent’s existing beliefs are quite distant from $\varphi$, so friction is higher.

To weight the evidence differently in order to  model uncertainty, one could add coefficients (e.g. $\lambda_1\mathbf{v}_\varphi + \lambda_2\mathbf{v}_E$) or use other similarity measures. The core idea is that "alignment" = "similarity" between the combined proposition/evidence vector and the agent’s belief vector.

In the previous section, we have characterized friction $F(\varphi,B,E)$ as  proportional to "misalignment." That is,
\enumsentence{$
     F(\varphi,B,E) \;\propto\; 1 - \text{alignment}(\varphi,B,E).
   $
   }
\noindent   So when alignment is high, friction is low, and vice versa.
We use the term "orthogonal" to indicate that the new proposition is "hard to assimilate." Orthogonality in vector spaces (cosine near zero) naturally corresponds to low alignment.

In both the set‐theoretic and vector‐based versions, one can incorporate $E$ to reflect how evidence changes the "effective proposition." More (or stronger) evidence typically boosts alignment with $B$, reducing friction.

Friction in epistemic updates occurs when new evidence $[E]\varphi$ conflicts with or is near-orthogonal to the agent's current belief state $[B]\neg\varphi$. 
Given the evidence-based DEL framework from the previous section, we can assume that friction modifies how propositions transition between the different banks. 
The transition rules from bank to bank can be viewed as follows:  
\enumsentence{
a. $\text{QBank} \xrightarrow{E \text{ sufficient, } F \text{ low}} \text{EBank}$;\\
b.  $\text{EBank} \xrightarrow{F \text{ near-zero}} \text{FBank}$.
}
 \subsection{Friction Equilibrium  in Discourse}

Dynamic Epistemic Friction (DEF) quantifies the resistance encountered during belief updates.  Our goal is to iteratively reduce friction in discourse in order to guide participants toward a better epistemic equilibrium. To this end, we assume:
\enumsentence{
a. $ \mathcal{D} = [\varphi_1, \varphi_2, \dots, \varphi_n] $: The set of propositions in the discourse;
\\
b.  $ \mathcal{S} = [B_1, B_2, \dots, B_m] $: The epistemic states of participants;\\
c.  $ E = [E_1, E_2 \dots E_n]$: Evidence associated with each proposition $ \varphi_i $.
}

\noindent We then proceed as follows:
\enumsentence{ \textbf{Initialize} the belief set:\\
Start with $ \mathcal{D}^0 = \mathcal{D} $ and $ \mathcal{S}^0 = \mathcal{S} $.\\
 Set iteration $ k = 0 $. This defines the basic elements required to measure friction and move toward equilibrium: the propositions discussed, the belief states of the participants, and the evidence supporting each proposition.

}

\enumsentence{ \textbf{Measure Friction:}
        For each proposition $ \phi_i \in \mathcal{D}^k $:    $
        F_i(a) = 1 - \text{alignment}(\phi_i, B_a, E_i)
        $,
        where $ F_i(a) $ is the friction for participant $ a $. Start the iterative equilibrium process from an initial state (no friction measured yet).
        Here, friction is measured by how misaligned each participant's belief state is with each proposition, given the available evidence: high alignment means that the participant's beliefs easily incorporate the proposition, resulting in low friction; low alignment means substantial disagreement or conflict, indicating high friction and a need for epistemic revision.
}
\enumsentence{\label{high-friction-corn-syrup} \textbf{Identify High-Friction Propositions:}
        For any $ a $, extract propositions $ \phi_i $ where $ F_i(a) > T $, the threshold for high friction. Let $ \mathcal{H} $ denote these high-friction propositions.  Such propositions are difficult for at least one participant to integrate into their beliefs, signaling a need for further discussion or clarification.
        
}
\enumsentence{ \textbf{Rank Propositions by Friction:}
        Rank $ \mathcal{H} $ by their average friction: 
        $$
        \text{Rank}(\phi_i) = \frac{1}{m} \sum_{a=1}^m F_i(a).
        $$
        Propositions are prioritized by how difficult (on average) they are to assimilate across all $m$ participants. Propositions with the highest average friction are candidates for clarification or refinement first, representing the greatest obstacle to achieving shared understanding.
}

\enumsentence{ \textbf{Refine High-Friction Propositions:}
        For the top-ranked $ \phi_j \in \mathcal{H}$, propose a refinement $ \phi_j^* $: \\
        (i) add evidence $ E_j' $, making the proposition easier to accept or (ii) modify $ \phi_j $ for better alignment with the current belief state.
 
}
\enumsentence{ \textbf{Update Belief States:}
\label{update}
        For each participant $a$:
        $$
        B_a^{k+1} = B_a^k + \Delta B_a,
        $$
        where $ \Delta B_a = - \nabla F(\phi_j^*, B_a, E_j') $. Beliefs are updated by applying a gradient step, effectively moving the belief states in a direction that reduces friction. The gradient descent step systematically adjusts participants' beliefs closer to propositions supported by evidence.
}
\enumsentence{ \textbf{Check Equilibrium:}
\label{equilibrium}
        Measure net friction:
        $$
        \mathcal{F}^k = \frac{1}{n \times m} \sum_{i=1}^n \sum_{a=1}^m F_i(a).
        $$
        If $ \mathcal{F}^k \leq T $, return equilibrium $ \mathcal{D}^k, \mathcal{S}^k $. This computes the net friction averaged across all propositions and participants, and quantifies how well the group is aligned as a whole. If the net friction is less than or equal to some threshold $T$, equilibrium is achieved. The participants' beliefs are now sufficiently aligned and no more substantial cognitive effort is required to maintain common ground.
}
\enumsentence{ \textbf{Iterate or Halt:}
        If $ k < \mu $, the maximum number of iterations, set $ k = k+1 $ and repeat. Otherwise, report no equilibrium.
}

\section{Empirical Demonstration}
\label{sec:empirical}

\begin{figure}
    \centering
    \includegraphics[width=\linewidth]{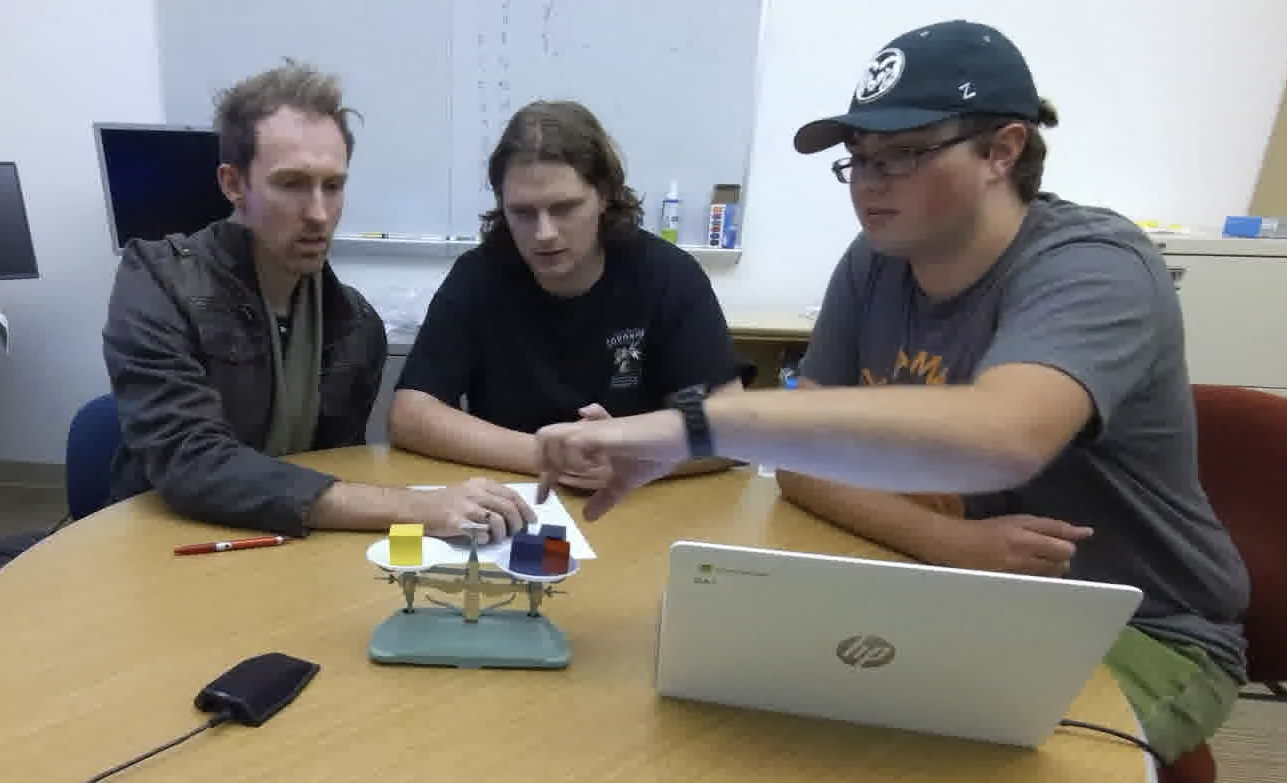}
    \caption{A group of 3 performing the Weights Task.}
    \label{fig:wtd}
\end{figure}

In this section we illustrate how the formal model detailed above can be operationalized to show how DEF can predict updates in the implied beliefs of real dialogue participants in a situated collaborative task. We experiment on the Weights Task Dataset (WTD;~\cite{khebour2024text}), in which triads collaborate to deduce the weights of differently-colored blocks using a balance scale (Fig.~\ref{fig:wtd}). The correct block weight assignments are $[red = 10g,blue = 10g,green = 20g,purple = 30g,yellow = 50g]$. 
The Weights Task is a collaborative task with one ideal convergent outcome. Use of this fixed condition allows the use of the aforementioned formal model in an analysis that can rigorously quantify the trajectory of convergence relative to a consistent ground truth and fit this model to the dynamics of any group, by controlling for the expected outcome while varying the individual participants.

We perform an experimental evaluation over 4 of of the 10 groups in the WTD, which are fully annotated with dialogue transcripts and the beliefs asserted by the three participants in each group \cite{vanderhoeven2025trace}, as indicated by speech, gesture, gaze, and action.  

For these analyses we adopt a simplified model of propositionalized belief states that can be used to construct multidimensional sparse vectors according to the assumptions given in (\ref{assumptions}), with a specific emphasis on {\it Saying is Believing}. Belief states are vectorized such that logical operators can be realized as arithmetic and algebraic operations, which gives intuitive properties like "alignment" and "irrelevance" analogies in measures like similarity and orthgonality (Sec.~\ref{sec:alignment}).

Given the 5 blocks in the task, belief states are vectorized in $\mathbb{R}^5$, ordered component-wise as in \citet{khebour2024text,khebour2024common} ($[red,blue,green,purple,yellow]$). Thus, an assertion of $red = 10 \wedge blue = 10$ is represented as $[10,10,0,0,0]$ indicating affirmative assertions regarding the weights of the red and blue blocks. The 0 components for the other blocks represent that no information regarding them is being asserted. Similarly, $green \neq 20$ would be represented as $[0,0,-20,0,0]$ (negative positioning toward $green = 20$, no other information asserted). Where blocks are related to other blocks by inequalities, the belief vector encodes a lower or upper bound regarding that block, such that $yellow < 40$ becomes $[0,0,0,0,40-\mathcal{U}(0,1)]$, thus anchoring the assertion relative to that weight value, in the appropriate direction.

\paragraph{Worked Example}

\begin{table}
\begin{tcolorbox}
\begin{dialogue}
    \speak{P1} Alright team, let's start weighing the blocks! Since we know the red block is 10 grams, should we weigh the blue block against it?
    \speak{P2} Great idea, I feel like the blue block will also be 10 grams. Let's do it!
    \speak{P1} \direct{Weighs red and blue blocks} They balance! So, the blue block is also 10 grams.
    \speak{P2} Confirmed! That's awesome. Now, what should we weigh next?
\end{dialogue}
\end{tcolorbox}
\caption{\label{tab:sampledialogue}Example generated dialogue.}
\end{table}

Consider the novel dialogue in Table~\ref{tab:sampledialogue}, generated using GPT-4 given a description of the task setup and goals. P1's assertion that the red and blue blocks both weigh 10 grams would be vectorized as $[10,10,0,0,0]$. Now consider a "frictive" utterance that pushes back on some of this assertion, which may inserted by an AI agent or another participant: "{\it Hey, let's not jump to conclusions about the blue block's weight just yet. What if it's not 10 grams?}" This assertion, expressing (conditionally) that $blue \neq 10$, would be vectorized as $[0,-10,0,0,0]$.

Now, letting $\vec\varphi_a$ be the focus participant's current belief vector, $\vec\varphi_b$ be the vector expressing the interlocutor's utterance, and $s = \text{sgn}(\vec\varphi_a \cdot \vec\varphi_b)$, consider an update operation akin to (\ref{update}):

\enumsentence{
    \label{update-vec}
    $\vec\varphi_a' = \vec\varphi_a + \text{min}\big(\beta,\alpha \times s\big)\times\text{CosSim}(\vec\varphi_a,\vec\varphi_b)\odot\vec\varphi_b$.
}

Here, we introduce some {\it friction coefficients} that allow us to tune how much empirical effect friction has on the belief update: $\alpha$ expresses how much "force" to apply the friction with (e.g., a scalar multiple of the gradient step), and $\beta$ establishes a "ceiling" on how much an assertion $\vec\varphi_b$ that is roughly aligned with $\vec\varphi_a$ can reinforce or "accelerate" it toward the status of an established belief, when compared to how much a contradictory or frictive assertion $\vec\varphi_b$ would suppress $\vec\varphi_a$.

Given the above $\vec\varphi_a = [10,10,0,0,0]$ and $\vec\varphi_b = [0,-10,0,0,0]$, with $\alpha = 1$ and $\beta = 1$, the updated belief state $\vec\varphi_a'$ after applying (\ref{update-vec}) becomes $[10,2.929,0,0,0]$. The assertion contradictory to $blue = 10$ renders it a "frictive" proposition and has lessened P1's epistemic commitment toward it. The precise component-wise values in the vector should not be taken to indicate what the participant believes the weight of the relevant block to be, but rather as an indicator of the {\it degree of belief} they have in the block's weight being the value assigned to it by the ground truth value assignment.

An interesting effect of these operations is that in certain circumstances when an assertion expresses information contrary to certain elements of the belief state but aligned with others, the effect may be greater on the component of the belief state against which friction is exercised. I.e., given $\vec\varphi_a = [10,10,20,0,0]$, $\vec\varphi_b = [10,-10,20,0,0]$, ($\vec\varphi_a$ and $\vec\varphi_b$ have the same red and green components but opposite blue components), $\alpha = 1$ and $\beta = 1$, the updated  $\vec\varphi_a' = [10,3.333,20,0,0]$, but given $\vec\varphi_a = [10,10,20,0,0]$, $\vec\varphi_b = [0,-10,20,0,0]$ (only the same green components, but opposite blue components), $\vec\varphi_a' = [10,4.523,20,0,0]$. That is, accordance on certain propositions gives differences more "weight" in the update.

\subsection{Experimental Procedure}

We adopt this procedure to evaluate the operationalization of our formal model of dynamic epistemic friction on the task of predicting what the final belief state (final state of FBank) of a target participant should be, given the utterances in the dialogue in order. This allows us to iteratively evaluate how the belief state evolves according to the DEF model. Since all groups in the Weights Task successfully deduced the weights of all blocks, the ground truth final state is a fixed $[10,10,20,30,50]$.

For tractability reasons we focus on modeling the only the belief state of the participant who speaks the {\it least} in each group.\footnote{Which specific participant this is may vary across groups and is not further explicated here.} This provides the greatest number of interlocutor utterances that affect the focus participant's belief state without updating it directly due to Saying is Believing (\ref{assumptions}).

As a consequence of Saying is Believing, if the focus participant makes a statement asserting a block weight or explicitly accepts another participant's positive and specific assertion about a block weight (e.g., $green = 20$, but not $green \neq 20$ or $green > blue$), then that value gets directly assigned to the relevant component in the focus participant's belief vector before the update function is run over subsequent utterances in the dialogue.

Under these conditions, we conduct the following procedure:

\begin{enumerate}
    \item Initialize the focus participant's "belief vector" from a uniform distribution $\mathcal{U}(0,10)$ and set the first ("red") element of the belief vector to 10. This reflects the initial state of the Weights Task where participants are told that the red block weighs 10g. The $\mathcal{U}(0,10)$ initialization ensures that belief vectors are not strictly 0 in most components, allowing for updates to actually shift the vector and reflecting participants' assumption (apparent in the original data) that weight values are somewhere in intervals of 10g.
    \item Iterate through the statements or acceptances of propositional content in each group dialogue. For each interlocutor utterance, appropriately encoded as described, apply the update function (\ref{update-vec}) to the focus participant's belief vector.
    \item At the end of each dialogue, extract the focus participant's final vectorized belief state.
    \item Fit a ridge regressor ($L_2$ scaling constant of 1) to map this extracted belief state to the ground truth final FBank $[10,10,20,30,50]$. We use a rotating leave-one-group-out split, such that we fit to the extracted final states from 3 of the annotated WTD dialogues and test on the remaining.
\end{enumerate}

This procedure allows us to test how well the final extracted belief state, as constructed using DEF interpretation of the naturally occurring friction in the dialogue, predicts the actual final FBank at the conclusion of the task. Due to inherent stochasticity in steps 1 and 4, we ran the aforementioned loop 100 times and average over the outcomes. We also conduct a variant where the belief state features extracted in step 3 include the concatenated final $k$ belief states in each dialogue, for $k \in \{1..4\}$. We use root mean-squared error for our primary metric, which puts error back in the original units and establishes how many "grams" the final prediction is off by in aggregate.

\subsection{Results}
\label{sec:results}

\begin{figure*}[t!]
    \centering
    \begin{subfigure}[b]{.45\textwidth}
        \centering
        \includegraphics[width=\linewidth]{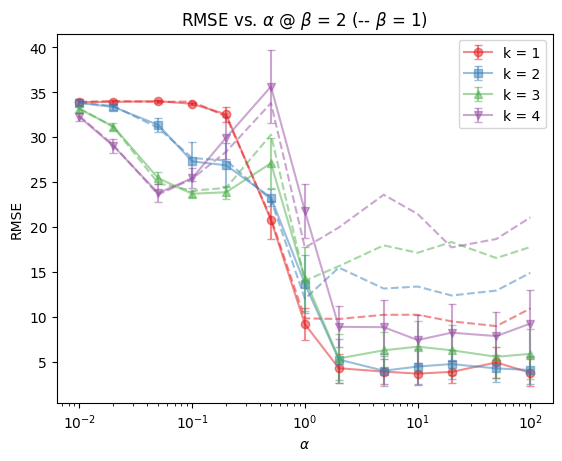}{}
        \caption{\label{fig:test-alpha}DEF peformance as a function of $\alpha$ with $\beta=2$ (dashed lines $\beta=1$ as a default baseline).}
    \end{subfigure}
    \hfill
    \begin{subfigure}[b]{.45\textwidth}
        \centering
        \includegraphics[width=\linewidth]{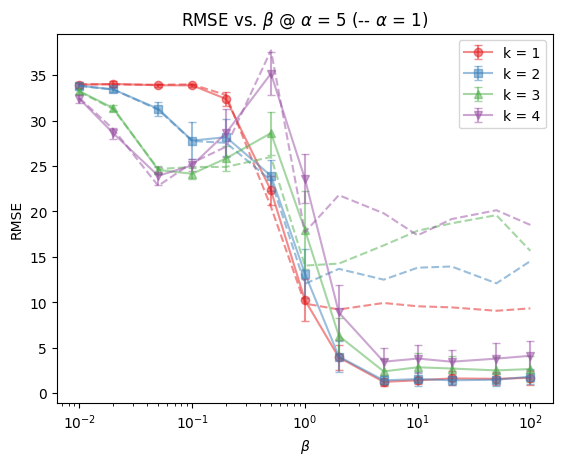}{}
        \caption{\label{fig:test-beta}DEF peformance as a function of $\beta$ with $\alpha=5$ (dashed lines $\alpha=1$ as a default baseline).}
    \end{subfigure}
    \caption{Effects of different $\alpha$ and $\beta$ values in the vector update function (\ref{update-vec}) on DEF peformance in FBank prediction. Values shown are averaged over leave-one-group-out cross-validation. Error bars represent standard error over 100 iterations, after cross-validation.}
    \label{fig:test-alpha-beta}
\end{figure*}

Table~\ref{tab:rmse} presents average weight prediction RMSE over each of the 4 test groups using 100 iterations of leave-one-group-out evaluation.

\begin{table}
    \centering
    \begin{tabular}{lcccc}\toprule
$k$ & \small {\bf Group 1}        & \small {\bf Group 2}      & \small {\bf Group 4}      & \small {\bf Group 5} \\\midrule
\small 1 & \small $2.613_{\pm 0.421}$   & \small $2.946_{\pm 0.595}$ & \small $7.678_{\pm 0.983}$ & \small $2.229_{\pm 0.233}$ \\
\small 2 & \small $1.889_{\pm 0.217}$   & \small $2.573_{\pm 0.395}$ & \small $10.731_{\pm 0.930}$ & \small $1.953_{\pm 0.204}$ \\
\small 3 & \small $4.449_{\pm 1.740}$   & \small $2.873_{\pm 0.631}$ & \small $13.292_{\pm 1.059}$ & \small $2.368_{\pm 0.453}$ \\
\small 4 & \small $5.187_{\pm 1.011}$   & \small $2.366_{\pm 0.505}$ & \small $17.501_{\pm 2.190}$ & \small $3.112_{\pm 1.012}$ \\
\bottomrule
\end{tabular}
    \caption{Average RMSE on weight prediction from DEF-constructed FBank over the 4 test groups, using update function (\ref{update-vec}) with friction coefficients $\alpha=5$ and $\beta=2$ and dialogue history length $k$.}
    \label{tab:rmse}
\end{table}

Using the FBank constructed with DEF, we are able to get very close to the true weight values, with an average RMSE of 2-3g for most test groups at low $k$, showing the efficacy of DEF in belief state tracking and prediction.\footnote{Group 4 is a shorter, sparser dialogue with fewer updates, and is therefore noisier.} This figure represents error across {\it all blocks} in all groups. In most cases the block weight introducing the most error into prediction was that of the yellow block. This is likely because the participants deduce the weight of the yellow (largest) block at the very end of the task, and while many utterances in the dialogue reiterate and deliberate upon the weights of the other blocks, fewer utterances discuss the yellow block, meaning there are fewer instances that shift the yellow component of the belief state vector toward the correct value. Thus, propositions pertaining to the yellow block, and $yellow = 50$ particularly, appear to be "high friction" propositions as in (\ref{high-friction-corn-syrup}) above.

Fig.~\ref{fig:test-alpha-beta} shows how the different values of friction coefficients $\alpha$ and $\beta$ as used in the vector update function (\ref{update-vec}) affect DEF's performance on belief state prediction. We performed a grid search through different values $\in \{0.01..100\}$ with a dialogue history window size $k$ of up to 4, using leave-one-group-out cross-validation. From this search, $\alpha=5$ and $\beta=2$ emerged as the best-performing combination.\footnote{These values were used to compute the groupwise results in Table~\ref{tab:rmse}.} Fig.~\ref{fig:test-alpha} presents RMSE as a function of $\alpha$ with $\beta$ fixed at 2, and Fig.~\ref{fig:test-beta} presents RMSE as a function of $\beta$ with $\alpha$ fixed at 5.

These figures show the importance of friction coefficients. $\alpha$ and $\beta$ are complementary and have similar effects, particularly as strong regularizers. When $k=4$, meaning longer dialogue history is used, prediction at lower $\alpha$ and $\beta$ values is noisy, with high RMSE and standard error. 
The lowest values of $\alpha$ and $\beta$ are effectively equivalent to a "minimal friction" or "no friction" setting in which interlocutor assertions are naively adopted by the listener.
However, as the friction coefficients grow larger, 
meaning more friction is effected by each update
, error drops dramatically. In other words, without enough friction, beliefs shift too rapidly toward ultimately incorrect positions. With too much, they become unchangeable and a slow trend of increasing error may emerge, particularly with longer dialogue histories and higher $\beta$. However, the right modulation of epistemic friction in the dialogue facilitates arriving at equilibrium as in (\ref{equilibrium}), where beliefs are guided toward agreed-upon propositions and remain there, achieving common ground.

Without consideration of epistemic friction, these propositions would naively be immediately adopted by every participant (cf. \citet{inan2025better}) A frictionless setting would not reflect the group dynamics of acceptance or refusal of propositions and would involve a greater error than all models involving friction.

\section{Conclusion}

In this paper, we presented a formal model of Dynamic Epistemic Friction (DEF) in dialogue, operationalized within the framework of Dynamic Epistemic Logic (DEL) and vector-based belief representations. We draw on the metaphor of friction as a physical force that changes the trajectory of a moving object as it encounters resistance and show that through the lens of DEL, analogous operations describe resistance to or accommodation of belief updates. Through empirical analyses using data from a situated collaborative task (the Weights Task Dataset), our results demonstrate that DEF effectively predicts participant belief updates by quantifying resistance encountered during belief revisions. Specifically, by operationalizing epistemic states and propositional assertions within an evidence-based dynamic logic with vector-based propositional encoding, we show that epistemic friction reliably indicates how smoothly participants integrate new evidence into their existing beliefs.

We should note that for a propositional vectorization as used in Sec.~\ref{sec:empirical} to hold, the vectorized propositional space needs to be at least roughly isotopic \cite{ethayarajh2019contextual,nath2023axomiyaberta}. This property is known to be at best inconsistent in modern LLMs \cite{machina2024anisotropy}, and for realistic data where the belief state may not be preannotated as in the WTD, a more sophisticated vectorization needs to be used such that arithmetic and algebraic operations have equivalent logical consequents. In order to retrieve high-quality vectorizations for realistic data, a vector-symbolic method~\cite{goldowsky2024analogical} could operate over a library of propositions. The extraction of these propositions from natural language or multimodal data is crucial for implementing dynamic epistemic friction in an end-to-end system  \cite{venkatesha2024propositional}.

While the direct application of LLMs on this task is underexplored, off-the-shelf LLMs are unlikely to be able to operationalize the {\it quantitative} formalism to belief revision outlined in this paper. \citet{inan2025better} show that friction improves qualitative mental modeling, and OTS systems may be able to provide qualitative judgments about belief at a given point in the task.  LLMs specifically aligned with a formal and functional definition of friction may be more adept at quantitative dialogue tasks. For instance, \citet{nath2025friction} show that LLMs optimized to be "friction agents" provide more effective interventions and guidance when optimized to be directly sensitive to "frictive states" (dialogue occurrences similar to how we define epistemic fricton here). \citet{AAMAS2025} propose three types of optimization strategies that exploit representations of group beliefs at various levels of depth. Following such lines as the above, combining DEF with qualitative judgments may allow for an even more accurate representation of human belief revision. Our theoretical formalism and empirical data are an important stepping-stone that shows the validity of this work in isolation, laying the foundation for further experimentation and implementation in end-to-end systems, especially those involving extraction from natural language.

Future work should investigate a vectorized approach to belief revision in adversarial or competitive tasks. These tasks may explicitly involve deception and other actions unobserved in the Weights Task; they may commonly use forms of communication prohibitive to collaborative settings but conducive to high performance in competitive environments. For example, the game of Diplomacy has been an object of study as a challenging setting for benchmarking communicative AI \cite{wongkamjan2024more}, and contains both cooperative and adversarial elements. The formalism of epistemic friction alloes an analysis of when convergent properties as seen in alliances, suddenly change character, as when former allies become adversaries, but one of the players may not realize this change has occurred. These state changes are accommodated by the DEF formalism, our experimental procedure can be used to detect these changes (in terms of changes in the convergent properties over time), and our experimental results provide a baseline convergent condition to test collaborator and adversary behavior against.

One could also use the common ground framework and dynamic epistemic friction updates to predict agent behavior. The difference between an agent's current belief state and the proposed updated state, as well as task history and agent behavior, can inform a classifier of how an agent might respond (immediate acceptance, counterargument, asking clarifying questions, etc.) to a given scenario at any moment. This analysis would show how deeply our model of dynamic epistemic friction corresponds to agent behavior and how it may serve as a necessary link in instructive or monitoring systems.

Our empirical results show DEF's effectiveness as a model and the importance of properly modulating the amount of friction in a dialogue (as shown as in the tuning of selected friction coefficients), but we did not compare DEF to other approaches as this novel model of friction in dialogue and novel method of evaluating does not have at present any direct competitors in the literature. It is not clear that existing methods, such as those used in the Dialogue State Tracking Challenge \cite{williams2016dialog} provide a meaningful comparison.

The friction metaphor serves as a bridge between logical updating operations (e.g., dynamic epistemic logic) and cognitively motivated geometric models (e.g., vector symbolic architectures). Importantly, it highlights the fact that belief change in dialogue is not always straightforward but can generate internal or inter-agent tension,  where the appropriate coefficient of friction plays a crucial role in mitigating misunderstanding. 

\section*{Acknowledgments}

This material is based in part upon work supported by Other Transaction award HR00112490377 from the U.S. Defense Advanced Research Projects Agency (DARPA) Friction for Accountability in Conversational Transactions (FACT) program. The views and conclusions contained in this document are those of the authors and should not be interpreted as representing the official policies, either expressed or implied, of the U.S. Government.

\bibliography{anthology,custom}

\end{document}